\documentclass[10pt,twocolumn,letterpaper]{article}

\usepackage{wacv}              

\usepackage{multirow}

\definecolor{wacvblue}{rgb}{0.21,0.49,0.74}
\usepackage[pagebackref,breaklinks,colorlinks,allcolors=wacvblue]{hyperref}

\def\wacvPaperID{2167} 
\def\confName{WACV}
\def\confYear{2026}

\title{SegMo: \underline{Seg}ment-aligned Text to 3D Human \underline{Mo}tion Generation}

\author{
Bowen Dang$^{1}$ \quad
Lin Wu$^{2}$ \quad
Xiaohang Yang$^{3}$ \quad
Zheng Yuan$^{1}$ \quad
Zhixiang Chen$^{1}$\thanks{Corresponding author}\\
$^{1}$University of Sheffield \quad
$^{2}$University of Glasgow \quad
$^{3}$Queen Mary University of London\\
{\footnotesize
\texttt{$^{1}$\{bdang2, zheng.yuan1, zhixiang.chen\}@sheffield.ac.uk} \quad
\texttt{$^{2}$l.wu.1@research.gla.ac.uk} \quad
\texttt{$^{3}$xiaohang.yang@qmul.ac.uk}}
}

\usepackage{tcolorbox}

\usepackage[accsupp]{axessibility}  

\begin{document}
\maketitle


\newcommand{\fmlRVQVAEEncoder}{
\begin{equation}
    V = \mathrm{Encoder}(M) \in \mathbb{R}^{n \times d},
\end{equation}
}

\newcommand{\fmlRVQVAEQuantization}{
\begin{equation}
    \tilde{V}_i = \sum_{j=0}^{k} \mathbf{z}_i^j,
\end{equation}
}

\newcommand{\fmlRVQVAEDecoder}{
\begin{equation}
    \hat{M} = \mathrm{Decoder}(\tilde{V}) \in \mathbb{R}^{N \times D}.
\end{equation}
}

\newcommand{\fmlMaskTransformer}{
\begin{equation}
    \mathbf{x}^0 = \mathrm{MaskTrans}(T, \mathbf{t}_{1:A}, \hat{\mathbf{x}^0}).
\end{equation}
}

\newcommand{\fmlResidualTransformer}{
\begin{equation}
    \mathbf{x}^i = \mathrm{ResTrans}(T, \mathbf{t}_{1:A}, \mathbf{x}^{0:i-1}, i).
\end{equation}
}


\newcommand{\fmlAgg}{
\begin{equation}
\begin{aligned}
    \mathbf{m}_i &= \mathrm{Agg}(\mathbf{x}^0, s_i, e_i) \\
                 &= \mathrm{MLP}(\mathrm{Concat}(\mathrm{mean}(\mathbf{x}_{s_i:e_i}^0), \max(\mathbf{x}_{s_i:e_i}^0))).
\end{aligned}
\end{equation}
}


\newcommand{\fmlMask}{
\begin{equation}
    \mathcal{L}_{mask} = -\sum_{i \in \mathcal{M}} \log p_{\theta} (\mathbf{x}_i^0 |\mathbf{c}, \hat{\mathbf{x}^0}).
\end{equation}
}

\newcommand{\fmlLoss}{
\begin{equation}
    \mathcal{L} = \mathcal{L}_{mask} + \lambda_{align} \mathcal{L}_{align}.
\end{equation}
}


\newcommand{\fmlN}{B \cdot A}


\newcommand{\fmlSample}{
\begin{equation}
\begin{aligned}
    \mathcal{L}_{t2m} &= -\frac{1}{\fmlN} \sum_{i=1}^{B} \sum_{j=1}^{A} \log \frac{\exp (\mathrm{sim}(\mathbf{t}_j^i, \mathbf{m}_j^i)/\tau)}{\sum_{k=1}^{A} \exp (\mathrm{sim}(\mathbf{t}_j^i, \mathbf{m}_k^i)/\tau)}, \\
    \mathcal{L}_{m2t} &= -\frac{1}{\fmlN} \sum_{i=1}^{B} \sum_{j=1}^{A} \log \frac{\exp (\mathrm{sim}(\mathbf{t}_j^i, \mathbf{m}_j^i)/\tau)}{\sum_{k=1}^{A} \exp (\mathrm{sim}(\mathbf{t}_k^i, \mathbf{m}_j^i)/\tau)}, \\
    \mathcal{L}_{align} &= \frac{1}{2}(\mathcal{L}_{t2m} + \mathcal{L}_{m2t}).
\end{aligned}
\end{equation}
}


\newcommand{\fmlBatchOne}{
\begin{equation}
\scalebox{0.9}{$
\begin{aligned}
    \mathcal{L}_{t2m} &= -\frac{1}{\fmlN} \sum_{i=1}^{B} \sum_{j=1}^{A} \log \frac{\exp (\mathrm{sim}(\mathbf{t}_j^i, \mathbf{m}_j^i)/\tau)}{\sum_{k=1}^{B} \sum_{l=1}^{A} \exp (\mathrm{sim}(\mathbf{t}_j^i, \mathbf{m}_l^k)/\tau)}, \\
    \mathcal{L}_{m2t} &= -\frac{1}{\fmlN} \sum_{i=1}^{B} \sum_{j=1}^{A} \log \frac{\exp (\mathrm{sim}(\mathbf{t}_j^i, \mathbf{m}_j^i)/\tau)}{\sum_{k=1}^{B} \sum_{l=1}^{A} \exp (\mathrm{sim}(\mathbf{t}_l^k, \mathbf{m}_j^i)/\tau)}, \\
    \mathcal{L}_{align} &= \frac{1}{2}(\mathcal{L}_{t2m} + \mathcal{L}_{m2t}).
\end{aligned}$}
\end{equation}
}




\newcommand{\fmlGlobal}{
\begin{equation}
\begin{aligned}
    \mathcal{L}_{t2m} &= -\frac{1}{B} \sum_{i=1}^{B} \log \frac{\exp (\mathrm{sim}(T^i, M^i)/\tau)}{\sum_{j=1}^{B} \exp (\mathrm{sim}(T^i, M^j)/\tau)}, \\
    \mathcal{L}_{m2t} &= -\frac{1}{B} \sum_{i=1}^{B} \log \frac{\exp (\mathrm{sim}(T^i, M^i)/\tau)}{\sum_{j=1}^{B} \exp (\mathrm{sim}(T^j, M^i)/\tau)}, \\
    \mathcal{L}_{align} &= \frac{1}{2}(\mathcal{L}_{t2m} + \mathcal{L}_{m2t}).
\end{aligned}
\end{equation}
}


\newcommand{\fmlToken}{
\begin{equation}
\begin{aligned}
    \mathcal{L}_{align} = - \frac{1}{B * L} \sum_{i=1}^{B} \sum_{j=1}^{L} \log \frac{\exp (\mathrm{sim}(\mathbf{t}_{seg(\mathbf{x}_j^i)}^i, \mathbf{x}_j^i)/\tau)}
    {\sum_{k=1}^{A} \exp (\mathrm{sim}(\mathbf{t}_k^i, \mathbf{x}_j^i)/\tau)}.
\end{aligned}
\end{equation}
}






\newcommand{\tabValue}[2]{#1$^{\pm #2}$}
\newcommand{\first}[1]{\textcolor{red}{#1}}
\newcommand{\second}[1]{\textcolor{blue}{#1}}
\newcommand{\ourName}{SegMo}


\newcommand{\tabRowReal}{
    Real &
    - &
    \tabValue{0.511}{.003} & 
    \tabValue{0.703}{.003} & 
    \tabValue{0.797}{.002} & 
    \tabValue{0.002}{.000} & 
    \tabValue{2.974}{.008} & 
    \tabValue{9.503}{.065} \\
}

\newcommand{\tabRowTMtoT}{
    TM2T \cite{TM2T} &
    GRU &
    \tabValue{0.424}{.003} & 
    \tabValue{0.618}{.003} & 
    \tabValue{0.729}{.002} & 
    \tabValue{1.501}{.017} & 
    \tabValue{3.467}{.011} & 
    \tabValue{8.589}{.076} \\
}

\newcommand{\tabRowTtoM}{
    T2M \cite{T2M} &
    GRU &
    \tabValue{0.455}{.003} & 
    \tabValue{0.636}{.003} & 
    \tabValue{0.736}{.002} & 
    \tabValue{1.087}{.021} & 
    \tabValue{3.347}{.008} & 
    \tabValue{9.175}{.083} \\
}

\newcommand{\tabRowMDM}{
    MDM \cite{MDM} & 
    CLIP &
    -- & 
    -- & 
    \tabValue{0.611}{.007} & 
    \tabValue{0.544}{.044} & 
    \tabValue{5.566}{.027} & 
    \first{\tabValue{9.559}{.086}} \\
}

\newcommand{\tabRowMotionDiffuse}{
    MotionDiffuse \cite{MotionDiffuse} & 
    Transformer &
    \tabValue{0.491}{.001} & 
    \tabValue{0.681}{.001} & 
    \tabValue{0.782}{.001} & 
    \tabValue{0.630}{.001} & 
    \tabValue{3.113}{.001} & 
    \tabValue{9.410}{.049} \\
}

\newcommand{\tabRowMLD}{
    MLD \cite{MLD} & 
    CLIP &
    \tabValue{0.481}{.003} & 
    \tabValue{0.673}{.003} & 
    \tabValue{0.772}{.002} & 
    \tabValue{0.473}{.013} & 
    \tabValue{3.196}{.010} & 
    \tabValue{9.724}{.082} \\
}

\newcommand{\tabRowTtoMGPT}{
    T2M-GPT \cite{T2M-GPT} & 
    CLIP &
    \tabValue{0.491}{.003} & 
    \tabValue{0.680}{.003} & 
    \tabValue{0.775}{.002} & 
    \tabValue{0.116}{.004} & 
    \tabValue{3.118}{.011} & 
    \tabValue{9.761}{.081} \\ 
}

\newcommand{\tabRowReMoDiffuse}{
    ReMoDiffuse \cite{ReMoDiffuse} & 
    CLIP &
    \tabValue{0.510}{.005} & 
    \tabValue{0.698}{.006} & 
    \tabValue{0.795}{.004} & 
    \tabValue{0.103}{.004} & 
    \tabValue{2.974}{.016} & 
    \tabValue{9.018}{.075} \\
}

\newcommand{\tabRowMMM}{
    MMM \cite{MMM} & 
    CLIP &
    \tabValue{0.515}{.002} & 
    \tabValue{0.708}{.002} & 
    \tabValue{0.804}{.002} & 
    \tabValue{0.089}{.005} & 
    \tabValue{2.926}{.007} & 
    \tabValue{9.577}{.050} \\
}

\newcommand{\tabRowCoMo}{
    CoMo \cite{CoMo} & 
    CLIP &
    \tabValue{0.502}{.002} & 
    \tabValue{0.692}{.007} & 
    \tabValue{0.790}{.002} & 
    \tabValue{0.262}{.004} & 
    \tabValue{3.032}{.015} & 
    \tabValue{9.936}{.066} \\
}

\newcommand{\tabRowLAMP}{
    LAMP \cite{LAMP} & 
    LAMP &
    \first{\tabValue{0.557}{.003}} & 
    \first{\tabValue{0.751}{.002}} & 
    \first{\tabValue{0.843}{.001}} & 
    \first{\tabValue{0.032}{.002}} & 
    \first{\tabValue{2.759}{.007}} & 
    \second{\tabValue{9.571}{.069}} \\
}

\newcommand{\tabRowMoMask}{
    MoMask (Baseline) \cite{MoMask} & 
    CLIP &
    \tabValue{0.521}{.002} & 
    \tabValue{0.713}{.002} & 
    \tabValue{0.807}{.002} & 
    \tabValue{0.045}{.002} & 
    \tabValue{2.958}{.008} & 
    -- \\
}

\newcommand{\tabRowOurs}{
    \ourName{} (Ours) &
    CLIP &
    \second{\tabValue{0.553}{0.003}} &
    \second{\tabValue{0.748}{0.003}} &
    \second{\tabValue{0.838}{0.002}} &
    \second{\tabValue{0.042}{0.002}} &
    \second{\tabValue{2.782}{0.008}} &
    \tabValue{9.632}{0.079} \\
}


\newcommand{\tabRowRealDis}{
    Real &
    \tabValue{0.511}{.003} & 
    \tabValue{0.703}{.003} & 
    \tabValue{0.797}{.002} & 
    \tabValue{0.002}{.000} & 
    \tabValue{2.974}{.008} & 
    \tabValue{9.503}{.065} \\
}

\newcommand{\tabRowOursCPD}{
    \ourName{} (CPD-based Seg) & 
    \second{\tabValue{0.548}{0.002}} &
    \second{\tabValue{0.744}{0.002}} &
    \second{\tabValue{0.834}{0.002}} &
    \second{\tabValue{0.046}{0.002}} &
    \second{\tabValue{2.804}{0.008}} &
    \tabValue{9.602}{0.070} \\
}

\newcommand{\tabRowOursClustering}{
    \ourName{} (clustering-based Seg) & 
    \second{\tabValue{0.548}{0.002}} &
    \tabValue{0.741}{0.002} &
    \tabValue{0.833}{0.002} &
    \tabValue{0.051}{0.002} &
    \tabValue{2.811}{0.008} &
    \second{\tabValue{9.585}{0.071}} \\
}

\newcommand{\tabRowOursNo}{
    \ourName{} (No Align) & 
    \tabValue{0.538}{0.003} &
    \tabValue{0.732}{0.003} &
    \tabValue{0.825}{0.002} &
    \tabValue{0.065}{0.003} &
    \tabValue{2.863}{0.009} &
    \tabValue{9.636}{0.065} \\
}

\newcommand{\tabRowOursBatch}{
    \ourName{} (Batch-Level Align) & 
    \tabValue{0.546}{0.002} &
    \tabValue{0.741}{0.002} &
    \tabValue{0.831}{0.002} &
    \tabValue{0.056}{0.003} &
    \tabValue{2.817}{0.008} &
    \first{\tabValue{9.506}{0.081}} \\
}

\newcommand{\tabRowOursGlobal}{
    \ourName{} (Global Align) & 
    \tabValue{0.539}{0.002} &
    \tabValue{0.732}{0.002} &
    \tabValue{0.826}{0.003} &
    \tabValue{0.072}{0.003} &
    \tabValue{2.861}{0.007} &
    \tabValue{9.632}{0.090} \\
}

\newcommand{\tabRowOursToken}{
    \ourName{} (Token Align) & 
    \tabValue{0.536}{0.002} &
    \tabValue{0.730}{0.002} &
    \tabValue{0.824}{0.002} &
    \tabValue{0.056}{0.002} &
    \tabValue{2.870}{0.007} &
    \tabValue{9.654}{0.066} \\
}

\newcommand{\tabRowOursDis}{
    \ourName{} & 
    \first{\tabValue{0.553}{0.003}} &
    \first{\tabValue{0.748}{0.003}} &
    \first{\tabValue{0.838}{0.002}} &
    \first{\tabValue{0.042}{0.002}} &
    \first{\tabValue{2.782}{0.008}} &
    \tabValue{9.632}{0.079} \\
}


\newcommand{\tabRowOursMean}{
    \ourName{} (Mean Pooling Agg) & 
    \tabValue{0.548}{0.003} &
    \tabValue{0.743}{0.002} &
    \tabValue{0.833}{0.002} &
    \tabValue{0.065}{0.002} &
    \tabValue{2.812}{0.008} &
    \tabValue{9.583}{0.052} \\
}

\newcommand{\tabRowOursMax}{
    \ourName{} (Max Pooling Agg) & 
    \tabValue{0.541}{0.002} &
    \tabValue{0.738}{0.002} &
    \tabValue{0.830}{0.002} &
    \tabValue{0.057}{0.002} &
    \tabValue{2.834}{0.006} &
    \tabValue{9.611}{0.062} \\
}

\newcommand{\tabRowOursQuery}{
    \ourName{} (Query-to-Token CA Agg) & 
    \tabValue{0.549}{0.002} &
    \second{\tabValue{0.744}{0.002}} &
    \second{\tabValue{0.837}{0.002}} &
    \tabValue{0.052}{0.002} &
    \second{\tabValue{2.783}{0.006}} &
    \first{\tabValue{9.559}{0.120}} \\
}

\newcommand{\tabRowOursCLS}{
    \ourName{} (CLS-token SA Agg) & 
    \tabValue{0.545}{0.002} &
    \tabValue{0.742}{0.002} &
    \tabValue{0.834}{0.002} &
    \tabValue{0.051}{0.002} &
    \tabValue{2.813}{0.007} &
    \tabValue{9.591}{0.102} \\
}

\newcommand{\tabRowOursLlama}{
    \ourName{} (Llama 3 8B) & 
    \second{\tabValue{0.550}{0.003}} &
    \second{\tabValue{0.744}{0.003}} &
    \tabValue{0.835}{0.002} &
    \second{\tabValue{0.047}{0.002}} &
    \tabValue{2.795}{0.007} &
    \second{\tabValue{9.574}{0.068}} \\
}

\newcommand{\tabRowOusQwen}{
    \ourName{} (Qwen 2.5 7B) & 
    \tabValue{0.542}{0.003} &
    \tabValue{0.738}{0.003} &
    \tabValue{0.831}{0.003} &
    \tabValue{0.063}{0.002} &
    \tabValue{2.830}{0.007} &
    \tabValue{9.587}{0.089} \\
}


\newcommand{\tabRowRealKIT}{
    Real & 
    - &
    \tabValue{0.424}{.005} & 
    \tabValue{0.649}{.006} & 
    \tabValue{0.779}{.006} & 
    \tabValue{0.031}{.004} & 
    \tabValue{2.788}{.012} & 
    \tabValue{11.08}{.097} \\ 
}

\newcommand{\tabRowTMtoTKIT}{
    TM2T \cite{TM2T} & 
    GRU &
    \tabValue{0.280}{.005} & 
    \tabValue{0.463}{.006} & 
    \tabValue{0.587}{.005} & 
    \tabValue{3.599}{.153} & 
    \tabValue{4.591}{.026} & 
    \tabValue{9.473}{.117} \\
}

\newcommand{\tabRowTtoMKIT}{
    T2M \cite{T2M} & 
    GRU &
    \tabValue{0.361}{.005} & 
    \tabValue{0.559}{.007} & 
    \tabValue{0.681}{.007} & 
    \tabValue{3.022}{.107} & 
    \tabValue{3.488}{.028} & 
    \tabValue{10.72}{.145} \\
}

\newcommand{\tabRowMDMKIT}{
    MDM \cite{MDM} & 
    CLIP &
    -- & 
    -- & 
    \tabValue{0.396}{.004} & 
    \tabValue{0.497}{.021} & 
    \tabValue{9.191}{.022} & 
    \tabValue{10.847}{.109} \\
}

\newcommand{\tabRowMotionDiffuseKIT}{
    MotionDiffuse \cite{MotionDiffuse} & 
    Transformer &
    \tabValue{0.417}{.004} & 
    \tabValue{0.621}{.004} & 
    \tabValue{0.739}{.004} & 
    \tabValue{1.934}{.064} & 
    \tabValue{2.958}{.005} & 
    \first{\tabValue{11.10}{.143}} \\
}

\newcommand{\tabRowMLDKIT}{
    MLD \cite{MLD} & 
    CLIP &
    \tabValue{0.390}{.008} & 
    \tabValue{0.609}{.008} & 
    \tabValue{0.734}{.007} & 
    \tabValue{0.404}{.027} & 
    \tabValue{3.204}{.027} & 
    \tabValue{10.80}{.117} \\ 
}

\newcommand{\tabRowTtoMGPTKIT}{
    T2M-GPT \cite{T2M-GPT} & 
    CLIP &
    \tabValue{0.402}{.006} & 
    \tabValue{0.619}{.005} & 
    \tabValue{0.737}{.006} & 
    \tabValue{0.717}{.041} & 
    \tabValue{3.053}{.026} & 
    \tabValue{10.86}{.094} \\
}

\newcommand{\tabRowReMoDiffuseKIT}{
    ReMoDiffuse \cite{ReMoDiffuse} & 
    CLIP &
    \tabValue{0.427}{.014} & 
    \tabValue{0.641}{.004} & 
    \tabValue{0.765}{.055} & 
    \second{\tabValue{0.155}{.006}} & 
    \tabValue{2.814}{.012} & 
    \tabValue{10.80}{.105} \\
}

\newcommand{\tabRowMMMKIT}{
    MMM \cite{MMM} & 
    CLIP &
    \tabValue{0.404}{.005} & 
    \tabValue{0.621}{.005} & 
    \tabValue{0.744}{.004} & 
    \tabValue{0.316}{.028} & 
    \tabValue{2.977}{.019} & 
    \tabValue{10.910}{.101} \\
}

\newcommand{\tabRowCoMoKIT}{
    CoMo \cite{CoMo} & 
    CLIP &
    \tabValue{0.422}{.009} & 
    \tabValue{0.638}{.007} & 
    \tabValue{0.765}{.011} & 
    \tabValue{0.332}{.045} & 
    \tabValue{2.873}{.021} & 
    \second{\tabValue{10.95}{.196}} \\
}

\newcommand{\tabRowLAMPKIT}{
    LAMP \cite{LAMP} & 
    LAMP &
    \first{\tabValue{0.479}{.006}} & 
    \first{\tabValue{0.691}{.005}} & 
    \first{\tabValue{0.826}{.005}} & 
    \first{\tabValue{0.141}{.013}} & 
    \first{\tabValue{2.704}{.018}} & 
    \tabValue{10.929}{.101} \\
}

\newcommand{\tabRowMoMaskKIT}{
    MoMask (Basline) \cite{MoMask} & 
    CLIP &
    \tabValue{0.433}{.007} & 
    \tabValue{0.656}{.005} & 
    \tabValue{0.781}{.005} & 
    \tabValue{0.204}{.011} & 
    \tabValue{2.779}{.022} & 
    -- \\
}

\newcommand{\tabRowOursKIT}{
    \ourName{} (Ours) & 
    CLIP &
    \second{\tabValue{0.443}{0.005}} &
    \second{\tabValue{0.663}{0.006}} &
    \second{\tabValue{0.784}{0.005}} &
    \tabValue{0.163}{0.007} &
    \second{\tabValue{2.733}{0.012}} &
    \tabValue{10.797}{0.113} \\
}


\newcommand{\tabCompTtoM}{
\begin{table*}[ht]
\centering
\makebox[\textwidth]{
\scalebox{0.85}{
    \begin{tabular}{lccccccc}
    \toprule
        \multirow{2}{*}{\textbf{Method}} 
        & \multirow{2}{*}{\textbf{Text Encoder}} 
        & \multicolumn{3}{c}{\textbf{R-Precision} $\uparrow$} 
        & \multirow{2}{*}{\textbf{FID} $\downarrow$} 
        & \multirow{2}{*}{\textbf{MM-Dist} $\downarrow$} 
        & \multirow{2}{*}{\textbf{Diversity} $\rightarrow$} \\
    \cmidrule(lr){3-5}
        & & Top 1 & Top 2 & Top 3 & & & \\
    \midrule
        \tabRowReal
    \midrule
        \tabRowTMtoT
        \tabRowTtoM
        \tabRowMotionDiffuse
        \tabRowMDM
        \tabRowMLD
        \tabRowTtoMGPT
        \tabRowReMoDiffuse
        \tabRowMMM      
        \tabRowCoMo        
        \tabRowLAMP
    \midrule
        \tabRowMoMask
        \tabRowOurs
    \bottomrule
    \end{tabular}
}
}
\caption{Comparison of text-conditional human motion generation on the HumanML3D test set. For each metric, we repeat the evaluation 20 times and report the average with a 95\% confidence interval. \first{Red} and \second{Blue} indicate the best and the second-best results.}
\label{tab:comp_t2m}
\end{table*}
}

\newcommand{\tabCompKIT}{
\begin{table*}[ht]
\centering
\makebox[\textwidth]{
\scalebox{0.85}{
    \begin{tabular}{lccccccc}
    \toprule
        \multirow{2}{*}{\textbf{Method}} 
        & \multirow{2}{*}{\textbf{Text Encoder}} 
        & \multicolumn{3}{c}{\textbf{R-Precision} $\uparrow$} 
        & \multirow{2}{*}{\textbf{FID} $\downarrow$} 
        & \multirow{2}{*}{\textbf{MM-Dist} $\downarrow$} 
        & \multirow{2}{*}{\textbf{Diversity} $\rightarrow$} \\
    \cmidrule(lr){3-5}
        & & Top 1 & Top 2 & Top 3 & & & \\
    \midrule
        \tabRowRealKIT
    \midrule
        \tabRowTMtoTKIT
        \tabRowTtoMKIT
        \tabRowMotionDiffuseKIT
        \tabRowMDMKIT
        \tabRowMLDKIT
        \tabRowTtoMGPTKIT
        \tabRowReMoDiffuseKIT
        \tabRowMMMKIT
        \tabRowCoMoKIT
        \tabRowLAMPKIT
    \midrule
        \tabRowMoMaskKIT
        \tabRowOursKIT
    \bottomrule
    \end{tabular}
}
}
\caption{Comparison of text-conditional human motion generation on the KIT-ML test set. For each metric, we repeat the evaluation 20 times and report the average with a 95\% confidence interval. \first{Red} and \second{Blue} indicate the best and the second-best results.}
\label{tab:comp_kit}
\end{table*}
}

\newcommand{\tabDiscussion}{
\begin{table*}[ht]
\centering
\makebox[\textwidth]{
\scalebox{0.85}{
    \begin{tabular}{lcccccc}
    \toprule
        \multirow{2}{*}{\textbf{Method}} 
        & \multicolumn{3}{c}{\textbf{R-Precision} $\uparrow$} 
        & \multirow{2}{*}{\textbf{FID} $\downarrow$} 
        & \multirow{2}{*}{\textbf{MM-Dist} $\downarrow$} 
        & \multirow{2}{*}{\textbf{Diversity} $\rightarrow$} \\
    \cmidrule(lr){2-4}
        & Top 1 & Top 2 & Top 3 & & & \\
    \midrule
        \tabRowRealDis
    \midrule
        \tabRowOursCPD
        \tabRowOursClustering
    \midrule
        \tabRowOursNo
        \tabRowOursBatch
        \tabRowOursGlobal
        \tabRowOursToken
    \midrule
        \tabRowOursDis
    \bottomrule
    \end{tabular}
}
}
\caption{Ablation results of replacing the segmentation and the alignment module on the HumanML3D test set. For each metric, we repeat the evaluation 20 times and report the average with a 95\% confidence interval. \first{Red} and \second{Blue} indicate the best and the second-best results.}
\label{tab:discussion}
\end{table*}
}


\newcommand{\tabCompSeg}{
\begin{table}[ht]
\centering
\scalebox{0.8}{
    \begin{tabular}{lcc}
    \toprule
        & \textbf{Error (mean)}  
        & \textbf{Error (std)} \\
    \midrule
        uniform Seg & 3.61 & 3.01 \\
        CPD-based Seg & 2.95 & 3.46 \\
        clustering-based Seg & 2.97 & 4.04 \\
    \bottomrule
    \end{tabular}
}
\caption{Evaluation of different motion segmentation methods on the BABEL dataset.}
\label{tab:comp_seg}
\end{table}
}

\newcommand{\tabLLMAggDis}{
\begin{table*}[ht]
\centering
\makebox[\textwidth]{
\scalebox{0.85}{
    \begin{tabular}{lcccccc}
    \toprule
        \multirow{2}{*}{\textbf{Method}} 
        & \multicolumn{3}{c}{\textbf{R-Precision} $\uparrow$} 
        & \multirow{2}{*}{\textbf{FID} $\downarrow$} 
        & \multirow{2}{*}{\textbf{MM-Dist} $\downarrow$} 
        & \multirow{2}{*}{\textbf{Diversity} $\rightarrow$} \\
    \cmidrule(lr){2-4}
        & Top 1 & Top 2 & Top 3 & & & \\
    \midrule
        \tabRowRealDis
    \midrule
        \tabRowOursLlama
        \tabRowOusQwen
    \midrule
        \tabRowOursMean
        \tabRowOursMax
        \tabRowOursQuery
        \tabRowOursCLS
    \midrule
        \tabRowOursDis
    \bottomrule
    \end{tabular}
}
}
\caption{Ablation results of replacing the LLMs and the aggregation module on the HumanML3D test set. For each metric, we repeat the evaluation 20 times and report the average with a 95\% confidence interval. \first{Red} and \second{Blue} indicate the best and the second-best results.}
\label{tab:llm_agg_dis}
\end{table*}
}
\newcommand{\figIdea}{
\begin{figure}[t]
  \centering
  \includegraphics[width=\linewidth]{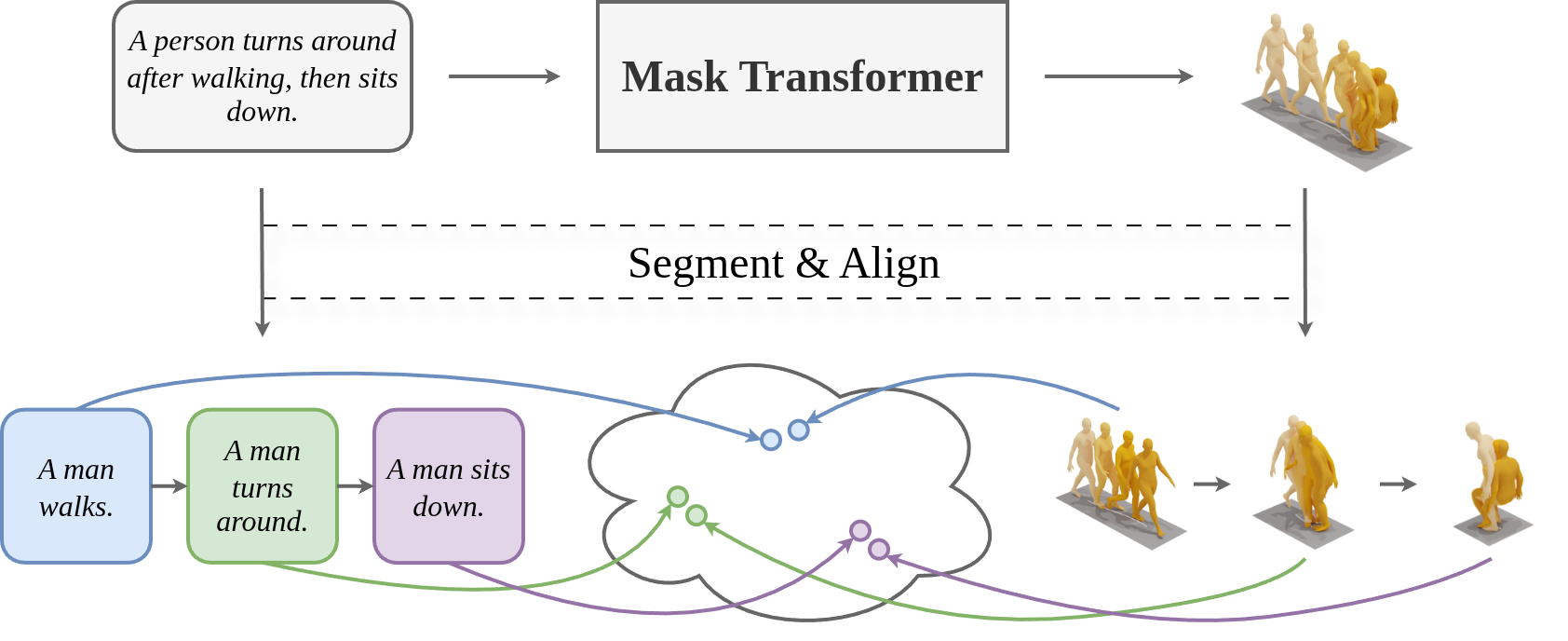}
  \caption{The main idea of our method. We decompose the complex motion description and motion sequence into simpler temporally ordered segments and align them in a shared embedding space to improve the accuracy and realism of generated motions.}
  \label{fig:idea}
\end{figure}
}

\newcommand{\figOverview}{
\begin{figure*}[t]
  \centering
  \includegraphics[width=0.95\linewidth]{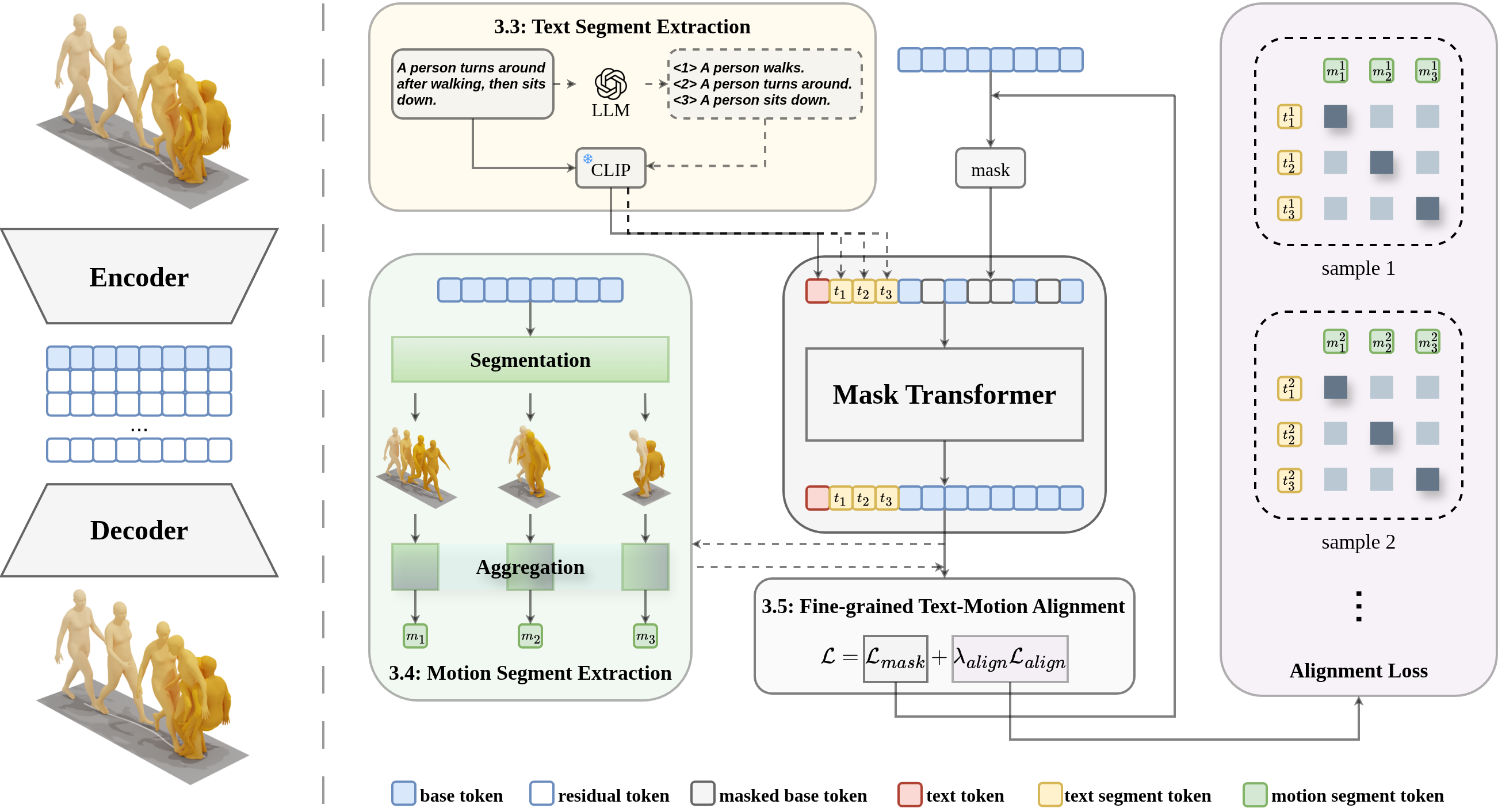}
  \caption{The overview of our method. \textbf{Left}: The Residual VQ-VAE encodes a continuous motion sequence into discrete motion tokens, including base tokens and residual tokens, which will be generated by the mask transformer and residual transformer, respectively. \textbf{Right}: The Mask Transformer predicts the masked base tokens conditioned on the textual description. To achieve segment-level fine-grained alignment, we introduce a \textit{Text Segment Extraction} module and a \textit{Motion Segment Extraction} module, which extract text and motion segments respectively, and align them through the \textit{Fine-grained Text-Motion Alignment} module.}
  \label{fig:overview}
\end{figure*}
}

\newcommand{\figQualitative}{
\begin{figure*}[t]
  \centering
  \includegraphics[width=0.95\linewidth]{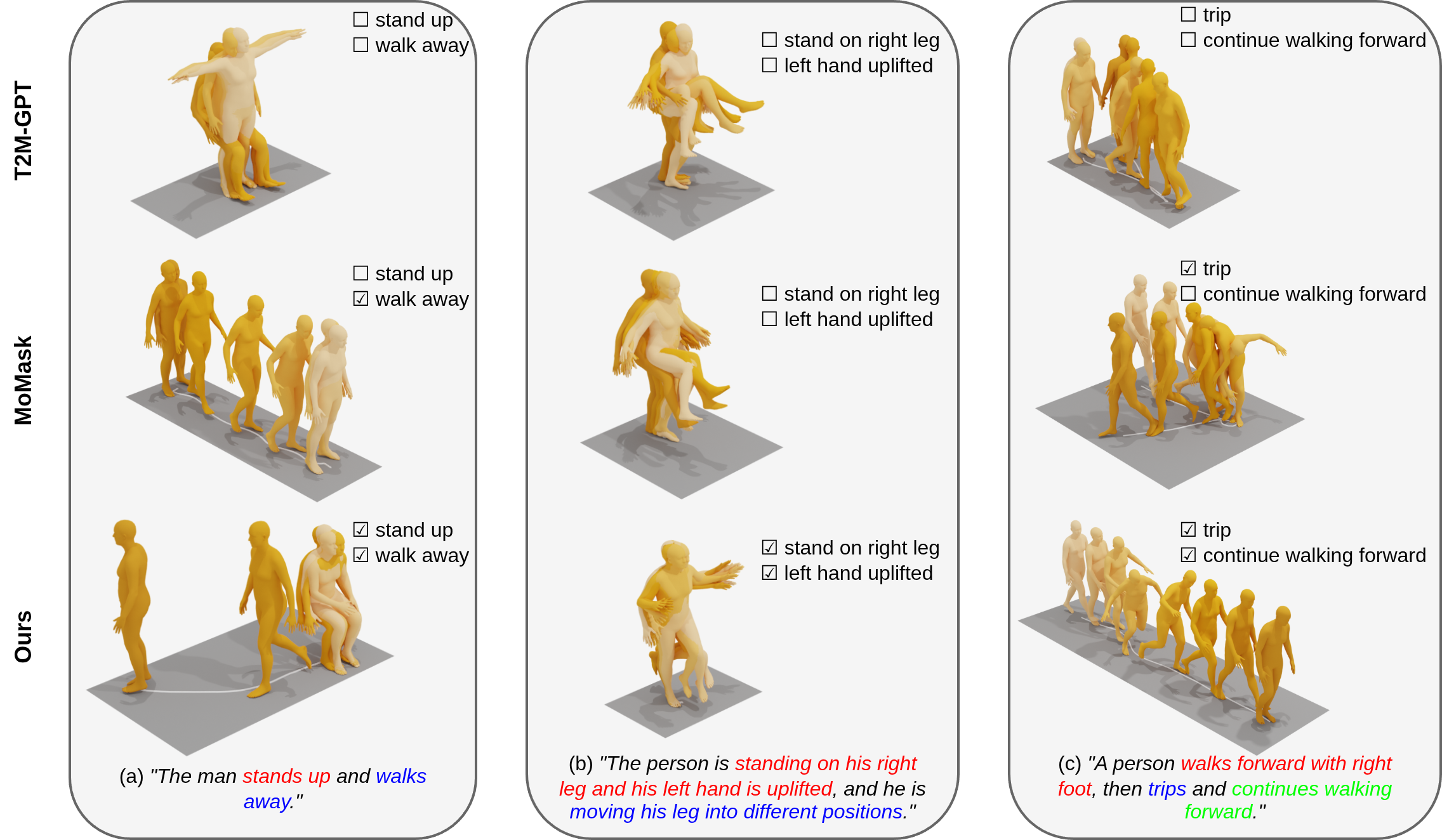}
  \caption{Qualitative comparison of T2M-GPT, MoMask, and Ours on the HumanML3D test set. A fixed number of keyframes is shown for each motion sequence. Please refer to the supplementary video for additional comparison results.}
  \label{fig:qualitative}
\end{figure*}
}

\newcommand{\figSegment}{
\begin{figure}[t]
  \centering
  \includegraphics[width=0.95\linewidth]{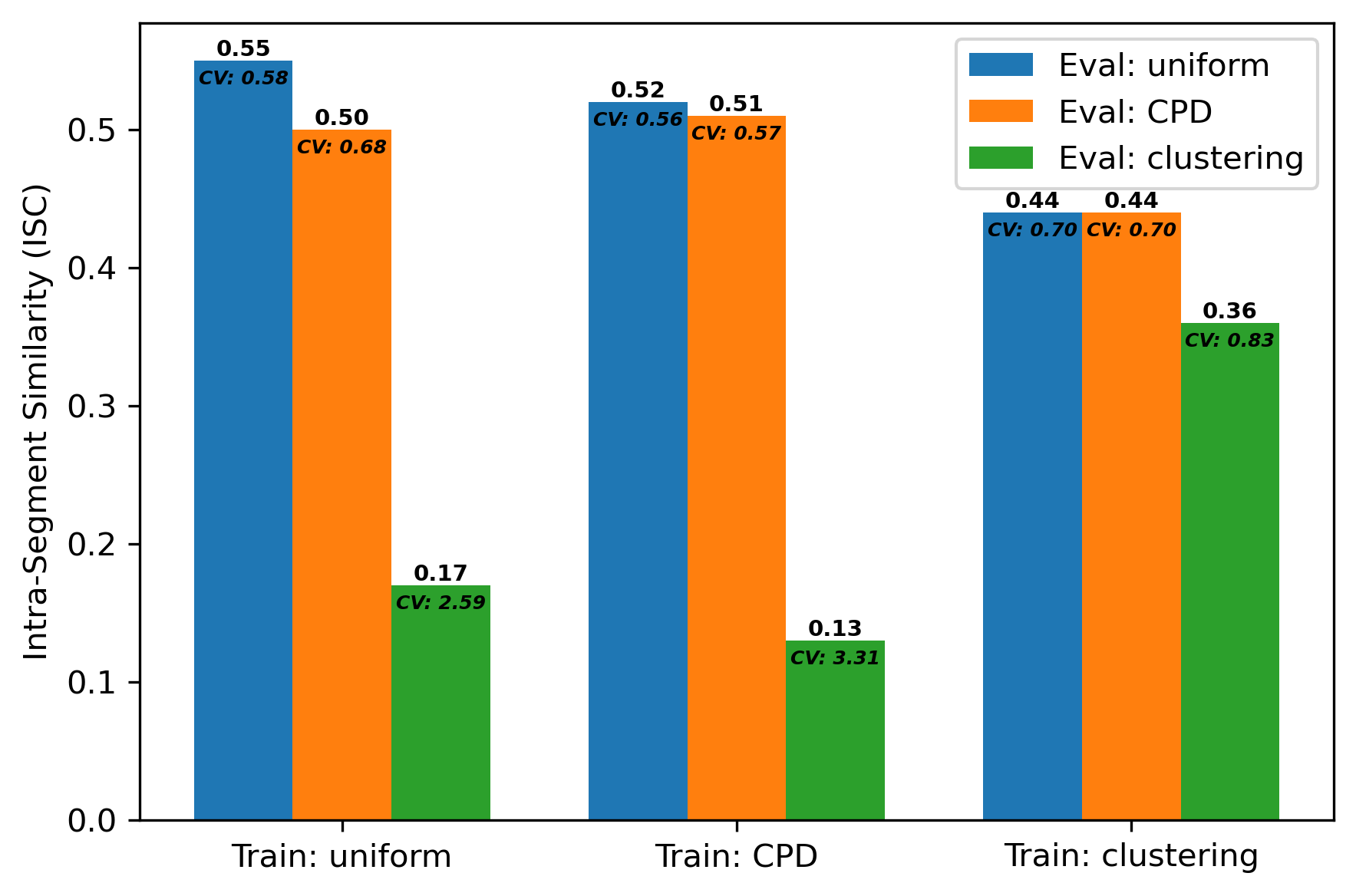}
  \caption{The Intra-Segment Consistency (ISC) of all models evaluated under different segmentation methods on the HumanML3D test set. The Coefficient of Variation (CV), defined as $\mathrm{std}(ISC)/\mathrm{mean}(ISC)$, is reported to assess stability. ``Train" denotes the segmentation method used for training, while ``Eval" denotes the segmentation method used for evaluation.} 
  \label{fig:segment}
\end{figure}
}

\newcommand{\figGrounding}{
\begin{figure}[htbp]
  \centering
  \includegraphics[width=\linewidth]{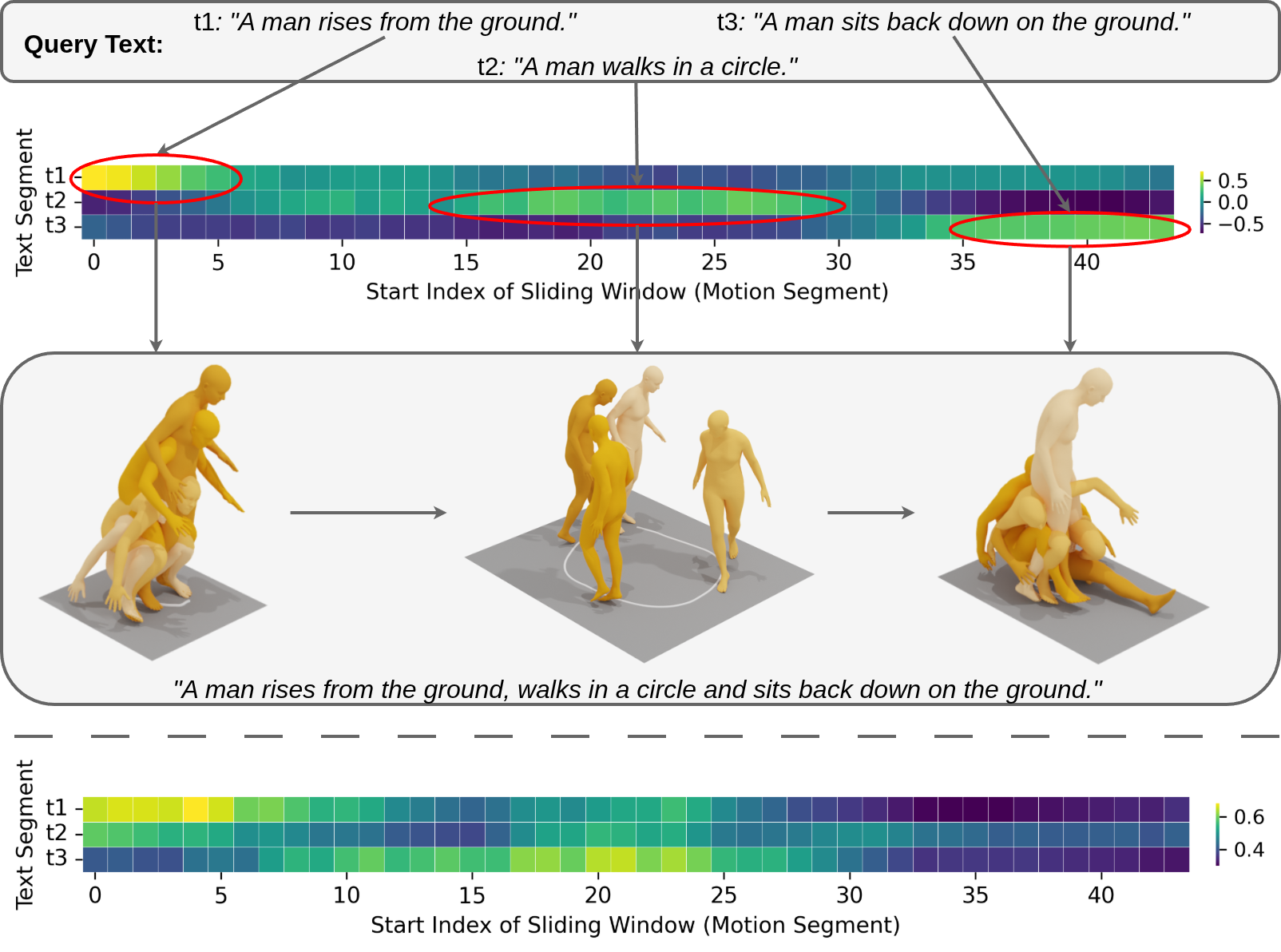}
  \caption{Example of motion grounding. \textbf{Top}: Results using the similarity map generated by our method. \textbf{Bottom}: Similarity map generated by MoMask~\cite{MoMask}. In each map, the x-axis denotes the start index of the sliding window and the y-axis denotes the text segment. The motion length is 49, and the window size is 5.}
  \label{fig:grounding}
\end{figure}
}


\newcommand{\figAggComp}{
\begin{figure}[htbp]
  \centering
  \includegraphics[width=\linewidth]{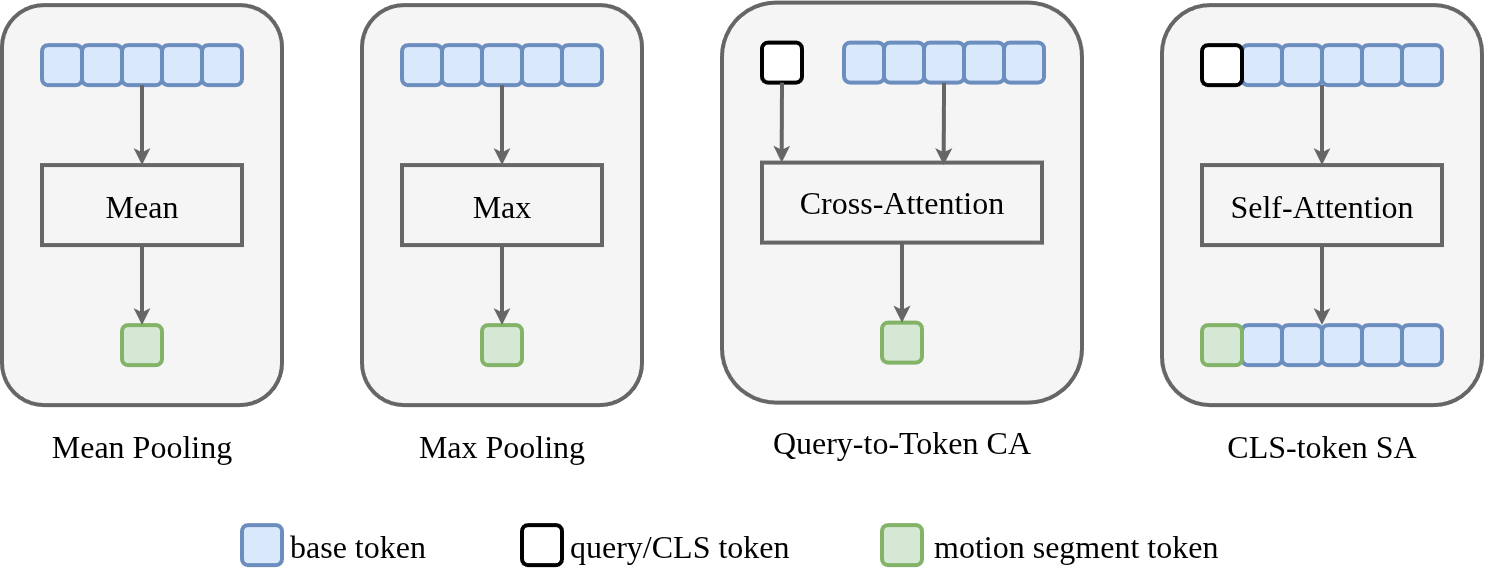}
  \caption{The comparison of different motion aggregation modules.}
  \label{fig:agg_comp}
\end{figure}
}

\newcommand{\figPrompt}{
\begin{figure*}[t]
    \centering
    \begin{tcolorbox}
    You are a helpful assistant. Your task is to extract and temporally order human actions from a sentence describing a person performing one or more actions.\\

    Guidelines:\\
    - If the sentence describes **only one action**, return it as a full sentence without using the ``\#" symbol.\\
    - If the sentence describes **multiple actions**, return each as a full sentence, in the correct temporal order, separated by the ``\#" symbol.\\
    - Preserve descriptive modifiers such as ``quickly", ``slowly", ``two times", ``as if", ``like", etc.\\
    - Normalize expressions such as ``appears to" or ``seems to" into direct action statements.\\
    - Do **not** include any extra text (e.g., ``Output:", quotes, or explanations). Return only the result string in the specified format.\\

    Examples:\\
    (1) Input: a person runs quickly after walking in a circle.\\
        Output: a person walks in a circle\#a person runs quickly.\\
        Comment: ``walks" comes before ``runs" due to the temporal cue ``after".\\

    (2) Input: a person jumps two times, then walks while waving the hands.\\
        Output: a person jumps two times\#a person walks while waving the hands.\\  
        Comment: ``two times" is a modifier; ``walks" and ``waves the hands" occur simultaneously.\\

    (3) Input: a person takes a box off the table and puts it on the floor.\\  
        Output: a person takes a box off the table\#a person puts a box on the floor.\\  
        Comment: The sentence contains two sequential actions—taking the box and then putting it down.\\

    (4) Input: a person is standing and waving the hands.\\  
        Output: a person is standing and waving the hands.\\ 
        Comment: The sentence contains two simultaneous actions—standing and waving the hands.\\

    (5) Input: a person is bowing left and right.\\  
        Output: a person is bowing left and right.\\  
        Comment: ``left and right" is a descriptive modifier and should be preserved.\\

    (6) Input: a person is jumping around like he is in an accident.\\  
        Output: a person is jumping around like he is in an accident.\\  
        Comment: ``like he is in an accident" is a descriptive modifier and should be preserved.\\

    (7) Input: a person appears to wave the hands.\\  
        Output: a person waves the hands.\\  
        Comment: ``appears to" is not an action and is removed.\\ 
        
    Now process the following input:
    \end{tcolorbox}
    \caption{Prompt for generating text segments using LLM.}
    \label{fig:prompt}
\end{figure*}
}

\newcommand{\figDistribution}{
\begin{figure*}[t]
  \centering
  \includegraphics[width=0.95\linewidth]{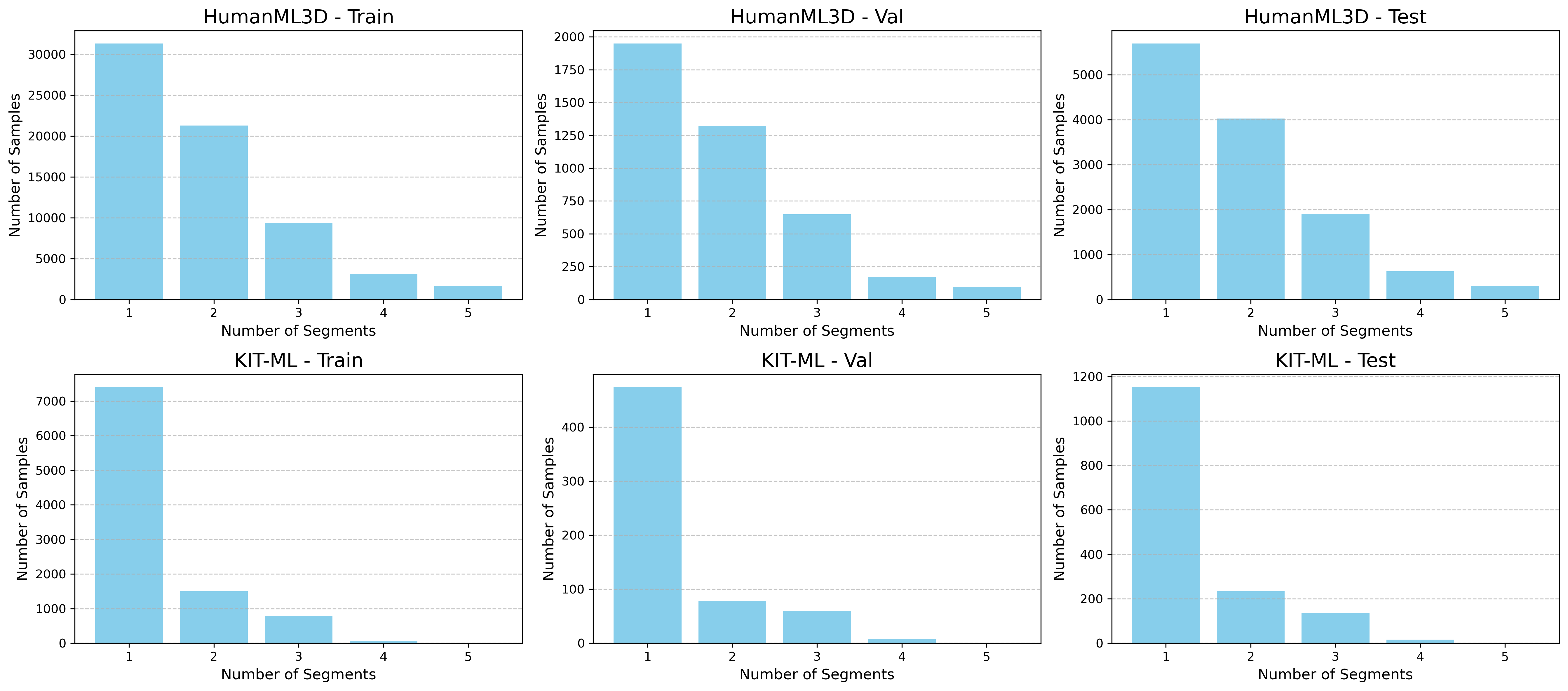}
  \caption{Segment number distributions of HumanML3D and KIT-ML datasets across training, validation, and test set. The x-axis denotes the number of segments, and the y-axis indicates the corresponding sample counts.}
  \label{fig:distribution}
\end{figure*}
}

\newcommand{\figFailure}{
\begin{figure}[htbp]
  \centering
  \includegraphics[width=\linewidth]{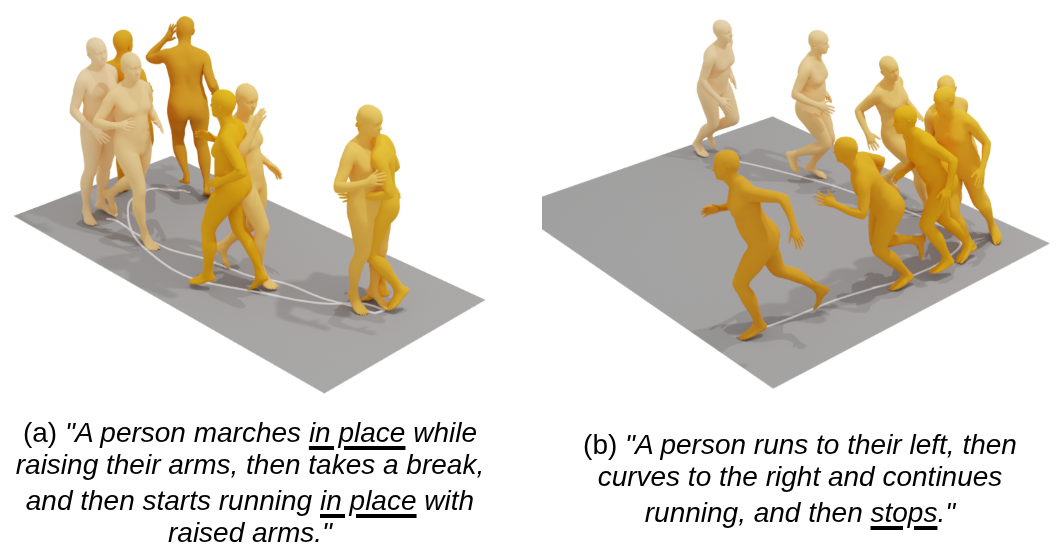}
  \caption{Failure cases of our method.}
  \label{fig:failure}
\end{figure}
}

\begin{abstract}
Generating 3D human motions from textual descriptions is an important research problem with broad applications in video games, virtual reality, and augmented reality. Recent methods align the textual description with human motion at the sequence level, neglecting the internal semantic structure of modalities. However, both motion descriptions and motion sequences can be naturally decomposed into smaller and semantically coherent segments, which can serve as atomic alignment units to achieve finer-grained correspondence. Motivated by this, we propose SegMo, a novel \textbf{Seg}ment-aligned text-conditioned human \textbf{Mo}tion generation framework to achieve fine-grained text–motion alignment. Our framework consists of three modules: (1) Text Segment Extraction, which decomposes complex textual descriptions into temporally ordered phrases, each representing a simple atomic action; (2) Motion Segment Extraction, which partitions complete motion sequences into corresponding motion segments; and (3) Fine-grained Text–Motion Alignment, which aligns text and motion segments with contrastive learning. Extensive experiments demonstrate that SegMo improves the strong baseline on two widely used datasets, achieving an improved TOP 1 score of 0.553 on the HumanML3D test set. Moreover, thanks to the learned shared embedding space for text and motion segments, SegMo can also be applied to retrieval-style tasks such as motion grounding and motion-to-text retrieval.
\end{abstract}
\section{Introduction}

Realistic human motion is important for applications in computer graphics, robotics, and immersive technologies such as video games, film production, virtual reality (VR), and augmented reality (AR). However, large-scale motion data mainly relies on costly motion capture systems with limited coverage, making it hard to represent the full range of human activities. To overcome these limitations, recent research has turned to motion generation~\cite{HMG-Survey}, where text-conditioned methods are gaining attention due to the intuitive and flexible control offered by natural language~\cite{T2M-Survey}.

\figIdea

To achieve text-conditioned motion generation, current methods typically rely on the CLIP~\cite{CLIP} text encoder to extract features from textual descriptions and use these features as conditions for the generation model. These methods typically pair an entire motion sequence with a single textual description, performing alignment only at the \textit{sequence level}. However, such alignment is inherently coarse because it \textit{neglects the internal structure of both modalities}. Consequently, models may produce generation errors such as missing actions, repeated actions, or incorrect action orders, which limit the accuracy and fidelity of synthesized motions.

To address this problem, we introduce a smaller alignment unit that enables more fine-grained text–motion alignment, thereby enhancing the accuracy and realism of generated motions. In practice, both motion descriptions and motion sequences can be decomposed into smaller, temporally and semantically coherent segments. In motion descriptions, \textit{a sentence is composed of phrases each describing an atomic motion}; in motion sequences, \textit{a sequence can be perceived as distinct actions connected by smooth transitions}. Cognitive studies, such as Event Segmentation Theory~\cite{EventSeg}, further support this view by showing that humans naturally segment continuous streams into meaningful events to facilitate perception and understanding. As illustrated in Figure~\ref{fig:idea}, a novel segment-level alignment constraint is imposed on the motion generation model to enable fine-grained text–motion correspondence and improve generation quality.



To achieve segment-level alignment, we first need to get the text and motion segments. On the text side, we leverage large language models (LLMs) to decompose complex textual descriptions into temporally ordered and semantically meaningful phrases. On the motion side, we explore several segmentation strategies, including uniform segmentation, Change-Point-Detection (CPD)-based segmentation, and clustering-based segmentation~\cite{KMeans}. Building on these components, we further introduce a fine-grained text–motion alignment module which performs contrastive learning within each sample to achieve more precise cross-modal correspondence. To the best of our knowledge, this is the first work to incorporate segment-level alignment into text-conditioned human motion generation. Extensive experiments demonstrate that our method generates more accurate and natural human motions than existing approaches. Moreover, the contrastive learning framework enables our method to generalize to retrieval-style tasks, such as motion grounding and motion-to-text retrieval.

Our main contributions can be summarized as follows:

\begin{itemize}
    \item We introduce a novel framework that decomposes both motion descriptions and motion sequences into atomic alignment units and performs fine-grained segment-level alignment via contrastive learning, enabling more accurate and natural motion generation.
    \item By learning a shared embedding for text and motion segments, our method naturally supports fine-grained retrieval-style tasks, including motion grounding and motion-to-text retrieval.
    \item Extensive experiments show that our method consistently improves upon strong baselines and achieves results comparable to the state-of-the-art without relying on the language-motion pretraining model.
\end{itemize}
\section{Related Work}

\figOverview

\paragraph{Text-Guided Human Motion Generation:} 
Existing human motion generation methods generally follow two paradigms: discrete models and continuous models~\cite{DisCoRD}.  


Discrete models treat motion sequences as discrete tokens. Typically, they adopt a two-stage generation process: first, a quantization model such as the Vector Quantized Variational Autoencoder (VQ-VAE)~\cite{VQ-VAE} discretizes continuous motion into a sequence of tokens; then, a generation model generates motion token sequences conditioned on the input text. Inspired by the success of GPT~\cite{GPT}, T2M-GPT~\cite{T2M-GPT} uses causal attention to generate motion autoregressively. However, autoregressive models are unidirectional and cannot capture bidirectional dependencies in motion data; they also require increased time for inference. Inspired by the masked modeling strategy in BERT-style models~\cite{BERT, MaskGIT}, some works adopt a transformer decoder structure and achieve promising results. For example, MMM~\cite{MMM} applies masked modeling to support motion editing tasks such as in-betweening, upper-body modification, and long-sequence generation. MoMask~\cite{MoMask} introduces a motion residual VQ-VAE~\cite{RVQ-VAE} to reduce quantization error. Accordingly, a mask transformer and a residual transformer are proposed to generate base and residual tokens, respectively. Building on MoMask, we propose to impose a segment-level alignment constraint to achieve more accurate and natural motion generation.  

In contrast to discrete models, which operate in the discrete motion space, continuous models directly model motion in continuous space using diffusion models. These models generate motion by gradually denoising an initial random signal into a realistic sequence through a learned conditional denoising process. MotionDiffuse~\cite{MotionDiffuse} is the first diffusion-based model for text-guided human motion generation. MDM~\cite{MDM} proposes predicting the sample directly instead of noise, enabling the use of geometric losses on motion positions and velocities. MLD~\cite{MLD} introduces a latent diffusion model that operates in a latent space learned via a Variational Autoencoder (VAE)~\cite{VAE}. MARDM~\cite{MARDM} revisits diffusion-based motion generation by addressing redundant representations and introducing a masked autoregressive framework with improved performance.


\paragraph{Text-Motion Alignment:} 
Text–motion alignment is a critical aspect of evaluating text-guided human motion generation. To ensure that the generated motion accurately reflects the textual description, most existing methods directly use the CLIP~\cite{CLIP} text encoder to extract text features as conditioning signals. Some approaches exploit more advanced text representations for fine-grained control. GraphMotion~\cite{GraphMotion} constructs a hierarchical semantic graph from textual descriptions and applies coarse-to-fine diffusion based on the graph. CoMo~\cite{CoMo} leverages LLMs to generate body-part-specific descriptions, enhancing the model's controllability.  

Inspired by contrastive learning in text–image modeling, several methods adopt similar ideas to improve text–motion alignment. MotionCLIP~\cite{MotionCLIP} aligns motions with the CLIP latent space to obtain semantically rich embeddings that support high-quality text-to-motion generation, editing, and recognition. MotionPatches~\cite{MotionPatches} applies contrastive learning to train a motion-language model, enabling both text-to-motion and motion-to-text retrieval. CAR~\cite{CAR} similarly decomposes textual descriptions into multiple phrases, but only for data augmentation in retrieval tasks. In contrast, we incorporate segment-level alignment into the motion generation task, leading to more accurate and natural motions.
\section{Method}

\subsection{Overview}

Our task is to generate 3D human motions from textual descriptions. Given a sentence describing a person performing certain actions, the goal is to produce the corresponding motion sequence that accurately reflects the input while remaining natural and realistic.

An overview of our method is shown in Figure~\ref{fig:overview}. We adopt MoMask~\cite{MoMask} as our baseline, which consists of two components: (1) a residual VQ-VAE (RVQ-VAE)~\cite{VQ-VAE, RVQ-VAE} that encodes the continuous motion sequence into discrete motion tokens, including tokens and residual tokens, and (2) a mask transformer and a residual transformer that generate the base and residual tokens, respectively. During generation, these tokens are fed into the RVQ-VAE decoder to reconstruct the motion sequence.

To enable fine-grained alignment, we define text segments and motion segments as the basic alignment units. A text segment is a standalone phrase that forms part of the full input text, while a motion segment denotes the corresponding sub-sequence of the motion. For example, in Figure~\ref{fig:overview}, the sentence \textit{``A person turns around after walking, then sits down."} can be decomposed into three temporally ordered text segments: [\textit{``A person walks."}, \textit{``A person turns around."}, \textit{``A person sits down."}], each corresponding to a sub-sequence of the motion.

Our main modification focuses on the mask transformer, where we introduce a \textit{Text Segment Extraction} module (Section~\ref{subsec:text_seg_ext}) and a \textit{Motion Segment Extraction} module (Section~\ref{subsec:motion_seg_ext}), which extract the text and motion segments, respectively. These segments are then aligned through a \textit{Fine-grained Text–Motion Alignment} module (Section~\ref{subsec:text_motion_align}), enabling segment-aligned motion generation.



\subsection{Model Structure}

\paragraph{Residual VQ-VAE:} 
Given an input continuous motion sequence $M \in \mathbb{R}^{N \times D}$, where $N$ is the number of frames and $D$ is the motion dimension, we first project it into the latent space using a 1D convolutional encoder:
\fmlRVQVAEEncoder
where $n/N$ is the downsampling ratio and $d$ is the latent dimension. Each latent vector $V_i$ is then quantized using a residual quantization technique:
\fmlRVQVAEQuantization
where $\mathbf{z}_i^0$ comes from the base codebook, $\mathbf{z}_i^j$ for $j>0$ quantizes the residual $V_i - \sum_{l=0}^{j-1} \mathbf{z}_i^l$ coming from the residual codebook, and $k$ is the layer number. Finally, the quantized latent sequence is projected back to the motion space using a 1D convolutional decoder:
\fmlRVQVAEDecoder
By using residual VQ-VAE to encode the motion, we can represent the motion sequence using base token sequence $\mathbf{x}^0$ and residual token sequences $\mathbf{x}^{1:k}$.

\paragraph{Mask Transformer:} 
The mask transformer is used to generate the base tokens, where our alignment module is applied. The input to the mask transformer is the text token $T$, text segment tokens $\mathbf{t}_{1:A}$, and masked base tokens $\hat{\mathbf{x}^0}$. The output is the predicted base tokens $\mathbf{x}^0$:
\fmlMaskTransformer
$A$ denotes the maximum segment number. During generation, starting from a fully masked empty sequence, we iteratively predict the tokens under a cosine scheduler \cite{MaskGIT, MoMask}.

\paragraph{Residual Transformer:} 
The residual transformer is used to generate the residual tokens and has a similar structure to the mask transformer. The input of the residual transformer is the text token $T$, text segment tokens $\mathbf{t}_{1:A}$, first $i-1$ layer tokens $\mathbf{x}^{0:i-1}$ and target layer index $i$. The output is the $i$-th layer tokens:
\fmlResidualTransformer
During training, a random target layer $i$ is selected, and all the tokens in the preceding $i-1$ layers are embedded and summed up as the token embedding input. During generation, after generating the base tokens, the residual tokens are produced layer by layer.

\subsection{Text Segment Extraction}
\label{subsec:text_seg_ext}

The text segment extraction module leverages the power of LLMs to decompose the raw textual description into temporally ordered text segments. When designing the prompting strategy, we adhere to the following principles to ensure that the decomposed text segments not only preserve the original information but also provide temporal structure information, which facilitates segment-level alignment between text and motion.

\begin{itemize}
\item Temporal ordering: Segments should follow the order of actions, rather than the order of appearance in the raw text input. For example, \textit{``A person runs after walking."} is decomposed into [\textit{``A person walks."}, \textit{``A person runs."}].
\item Preserving details: Segments should retain important modifiers such as \textit{``slowly"}, \textit{``two times"}, or \textit{``as if + description"}. For instance, \textit{``A person runs quickly, then jumps two times."} is decomposed into [\textit{``A person runs quickly."}, \textit{``A person jumps two times."}].
\item Simultaneous actions: If multiple actions occur simultaneously, they are treated as a single segment. For example, \textit{``A person is standing and waving the hands."} remains as one segment.
\end{itemize}

Following these rules, the input textual description is parsed into multiple text segments using an LLM, where each segment corresponds to a short motion sequence. We tested different medium-size LLMs and found that Qwen 3:8B~\cite{Qwen3} produces reliable and consistent results. The results using different LLMs and the detailed prompt are provided in the supplementary material (Section A).

In practice, the processing step is conducted offline. After obtaining the text segments, we use the CLIP~\cite{CLIP} text encoder to extract features for both the raw textual description and its segments, resulting in one text token $T$ and up to $A$ text segment tokens $\mathbf{t}_{1:A}$. These features are then fed into both the mask transformer and residual transformer as the conditioning signals.

\subsection{Motion Segment Extraction}
\label{subsec:motion_seg_ext}

\tabCompTtoM

After obtaining the text segments of the textual description, the next step is to extract the corresponding motion segments, which will serve as the atomic alignment units.

The segmentation module partitions the motion into the same number of segments as the text. We adopt a uniform segmentation strategy as our default choice, where motion tokens are evenly distributed across the text segments, ensuring balanced segment lengths that facilitate stable text–motion alignment and consistent model training. Given the predicted base tokens $\mathbf{x}^0$ from the mask transformer, along with the start index $s_i$ and end index $e_i$ of the $i$-th motion segment from the segmentation module, the motion segment aggregation module outputs the corresponding motion segment $\mathbf{m}_i$:
\fmlAgg

For comparison, we also explored other feasible segmentation methods, including Change-Point-Detection (CPD)-based and Clustering-based segmentation approaches. Details are provided in Section~\ref{para:motion_seg_module}. While these methods can be applied, uniform segmentation consistently produces the most stable alignment and training performance, resulting in superior generation quality compared to alternative methods.

\subsection{Fine-Grained Text-Motion Alignment}
\label{subsec:text_motion_align}

After obtaining the text and motion segments, the next step is to align them in a shared embedding space. We design an alignment module to establish fine-grained correspondence between text and motion. Here, we introduce the loss function used to train the mask transformer under this segment-level alignment constraint.

\paragraph{Mask Loss:} 
We denote the text condition as $c$ and randomly mask a portion of the base tokens to obtain $\hat{\mathbf{x}^0}$. The mask loss minimizes the negative log-likelihood of the target predictions for the masked base tokens:
\fmlMask
Here, $\mathcal{M}$ denotes the index set of the masked base tokens.

\tabCompKIT

\paragraph{Alignment Loss:} 
As illustrated in Figure~\ref{fig:overview}, the alignment module ensures that each text segment is paired with its corresponding motion segment within a sample, avoiding semantic interference across different samples. This differs from CLIP-style implementations~\cite{CLIP}, which compute contrastive loss across all samples in a batch. The alignment loss is defined as follows:
\fmlSample
Here, $B$ denotes the batch size, $A$ denotes the maximum number of segments per sample, $\mathrm{sim}(\mathbf{t}_j^i, \mathbf{m}_k^i)$ denotes the cosine similarity between the $j$-th text segment and the $k$-th motion segment in the $i$-th sample, and $\tau$ denotes the temperature parameter. The alignment loss enforces the text/motion segment to be closer to the paired motion/text segment, while being far away from other motion/text segments. 

\paragraph{Overall Objective:} 
The total loss for training the mask Transformer combines the mask loss and alignment loss:
\fmlLoss
Here, $\lambda_{align}$ denotes the weight for the alignment loss.

\section{Experiments}

\subsection{Experiment Settings}


\paragraph{Datasets:} 
We conduct experiments on two widely used datasets: HumanML3D~\cite{T2M} and KIT Motion-Language (KIT-ML)~\cite{KIT-ML}. HumanML3D combines motion sequences from the HumanAct12~\cite{HumanAct12} and AMASS~\cite{AMASS} datasets, pairing each sequence with three distinct textual descriptions, resulting in 14.6k motion sequences and 44.9k textual annotations. KIT-ML is a smaller dataset containing 3.9k motion sequences and 6.3k textual annotations. The segment number distributions for both datasets are provided in the supplementary material (Section E). For both datasets, we adopt the pose representation from T2M~\cite{T2M}.

\paragraph{Evaluation Metrics:} 
We follow the standard evaluation protocols from T2M~\cite{T2M} to assess the motion generation model. The metrics are categorized into three groups: (1) R-Precision and Multimodal Distance (MM-Dist), which measure how well the generated motions align with the input textual descriptions. R-Precision is reported with Top-1, Top-2, and Top-3 accuracy. (2) Fréchet Inception Distance (FID), which evaluates the realism of the generated motions by computing the distribution distance between the generated and ground-truth motion features. (3) Diversity, which is calculated by averaging the Euclidean distances of 300 randomly sampled motion pairs. The closer this value is to that of real motions, the better. 

\paragraph{Implementation Details:} 
Our model is implemented in PyTorch and evaluated on an NVIDIA A100 GPU. We follow the same training setup as MoMask~\cite{MoMask} for the residual VQ-VAE, the mask transformer, and the residual transformer. We calculate the average segment number of both datasets and find that setting the maximum number of segments to 5 can cover most samples. The temperature for computing the alignment loss is fixed at 0.1, and the loss weight is set to 1.0 on the HumanML3D dataset and 0.1 on the KIT-ML dataset.

\figQualitative

\subsection{Evaluation}

\paragraph{Quantitative Comparison:} 
We evaluate our method against a range of existing approaches, including discrete models~\cite{TM2T, T2M, T2M-GPT, MMM, CoMo, LAMP, MoMask} and continuous models~\cite{MotionDiffuse, MDM, MLD, ReMoDiffuse}. The results on the HumanML3D dataset are presented in Table~\ref{tab:comp_t2m}. Compared to MoMask~\cite{MoMask}, a strong baseline, our approach achieves significant improvements, particularly on text-motion alignment metrics such as R-Precision and MM-Dist, while also maintaining or improving performance on other metrics. Furthermore, our method achieves performance comparable to LAMP~\cite{LAMP}, the current state-of-the-art (SOTA) approach, without relying on a language-motion pretraining model. The results on the KIT-ML dataset are shown in Table~\ref{tab:comp_kit}. Although most sequences in KIT-ML contain only a single action, our alignment module improves all metrics over the baseline method, while achieving a comparable FID to the SOTA method.

\paragraph{Qualitative Comparison:} 
Figure~\ref{fig:qualitative} presents a qualitative comparison of our approach with T2M-GPT~\cite{T2M-GPT} and MoMask. In Figure~\ref{fig:qualitative} (a), both T2M-GPT and MoMask start from a standing pose and fail to generate the \textit{``stand up"} action. In Figure~\ref{fig:qualitative} (b), existing methods fail to accurately capture or distinguish critical descriptors such as \textit{``left"} and \textit{``right"}, resulting in mismatched motions. In Figure~\ref{fig:qualitative} (c), the compared methods do not \textit{``continue walking forward"}, leading to incorrect motion segments. Overall, our method effectively captures segment-level details, generating motions that are both natural and consistent with the textual descriptions.

\subsection{Discussion}

Table~\ref{tab:discussion} presents an ablation study evaluating the impact of different design choices by replacing individual modules.

\paragraph{Motion Segmentation Module:} 
\label{para:motion_seg_module}



Human motion is continuous and often contains long transition phases, making action boundaries inherently ambiguous. This motivates exploring different segmentation strategies, including uniform, Change-Point-Detection (CPD)-based, and clustering-based approaches. Further details are reported in the supplementary material (Section B). As reported in Table~\ref{tab:discussion} and the supplementary material (Table 2), the mean segmentation error appears to have little effect on performance, since transition frames can plausibly belong to either neighboring action. In contrast, the variance of segmentation errors is more critical: \textit{high variance leads to inconsistent boundaries and noisy alignments, destabilizing optimization, whereas low variance produces more consistent pairs, enabling stable training and high-quality generation.} Among all methods, uniform segmentation naturally achieves the lowest error variance, yielding the most consistent improvements in generation quality, as shown in Table~\ref{tab:discussion}.

\figSegment

To further evaluate how segmentation affects text–motion alignment, we introduce the Intra-Segment Consistency (ISC) score, defined as the cosine similarity between paired text and motion segments in the learned embedding space. ISC reflects both segmentation quality and the model’s alignment ability, depending jointly on the trained model and the segmentation method. Figure~\ref{fig:segment} highlights two key observations: (1) all models achieve the highest ISC when evaluated with uniform segmentation, whereas clustering-based segmentation introduces high instability and reduces alignment quality; (2) when evaluated with the same segmentation method used during training, models trained with uniform segmentation achieve the highest ISC, consistent with the generation metrics.


\tabDiscussion

\paragraph{Text-Motion Alignment Module:} 
For the text-motion alignment module, we also evaluated different strategies: (1) Batch-Level Alignment, where text and motion segments from different samples within a batch are concatenated, and the alignment loss is computed within the batch; (2) Global Alignment, where the alignment loss is calculated between the entire textual description and the complete motion sequence within a batch; (3) Token Alignment, where the alignment loss is computed directly between motion tokens and their corresponding text segments within a sample. Further details are provided in the supplementary material (Section D). In summary, our proposed alignment module provides a more balanced design: it preserves fine-grained text–motion correspondences while maintaining sufficient semantic context, leading to more robust alignment and better overall performance.

\subsection{Applications}

Our alignment module enforces each text segment and its corresponding motion segment to be close in a shared embedding space via contrastive learning. This property naturally enables our model to extend beyond generation tasks to retrieval-style applications. In particular, we explore two tasks: motion grounding and motion-to-text retrieval.

\paragraph{Motion Grounding:} 
Motion grounding aims to localize the corresponding motion segment for a query text within a motion sequence~\cite{Grounding}. Specifically, we input both the query text and the motion sequence into our model to map them into the shared embedding space. Then, we compute the cosine similarity between the query text segment $\mathbf{t}_q$ and all candidate motion segments $\mathcal{M}$. The candidate motion segments can be obtained using a sliding window. The motion segment with the highest similarity is regarded as the localized result: $\mathbf{m}^* = \arg\max_{\mathbf{m} \in \mathcal{M}} \mathrm{sim}(\mathbf{t}_q, \mathbf{m})$. Figure~\ref{fig:grounding} illustrates an example of motion grounding, showing how our model effectively matches a query text with its corresponding motion segment, while the baseline fails to establish the correct correspondence.

\figGrounding

\paragraph{Motion-to-Text Retrieval:} 
In the reverse direction, our model can also perform motion-to-text retrieval, where the goal is to find the most matching text description for a given motion segment. Given a query motion segment $\mathbf{m}_q$, we retrieve the text segment with the highest similarity from the candidate set $\mathcal{T}$: $\mathbf{t}^* = \arg\max_{\mathbf{t} \in \mathcal{T}} \mathrm{sim}(\mathbf{t}, \mathbf{m}_q)$.

\section{Conclusion}

In this work, we introduce SegMo, a segment-aligned text-conditional human motion generation method. Our main idea is to minimize the alignment unit by decomposing complex motion descriptions and motion sequences into temporally ordered segments and aligning them in a shared embedding space. Evaluation on standard benchmarks shows that SegMo achieves fine-grained text–motion alignment, generating more accurate and natural human motions. 


\noindent \textbf{Limitation:} The limitation of our method lies in the motion segmentation module. While uniform segmentation is stable and outperforms other approaches, it is not precise enough for fine-grained alignment. In the future, we will improve segmentation accuracy and leverage datasets with detailed segmentation annotations to enhance the model’s alignment precision.




\section*{Acknowledgement}

This work made use of the facilities of the N8 Centre of Excellence in Computationally Intensive Research (N8 CIR) provided and funded by the N8 research partnership and EPSRC (Grant No. EP/T022167/1). The Centre is co-ordinated by the Universities of Durham, Manchester and York.

{
    \small
    \bibliographystyle{ieeenat_fullname}
    \bibliography{main}
}

\setcounter{section}{0}
\renewcommand{\thesection}{\Alph{section}}
\section{Text Segment Extraction Module}

\subsection{Ablation Results using Different LLMs}

We evaluated the impact of replacing the LLM used in the text segment extraction module by substituting Qwen 3:8B~\cite{Qwen3} with Llama 3:8B~\cite{Llama3} and Qwen 2.5:7B~\cite{Qwen2.5}, as reported in Table~\ref{tab:llm_agg_dis}. As shown, Qwen 3:8B produced higher-quality text segments, resulting in superior overall performance.

\tabLLMAggDis

\subsection{Prompt for Text Segment Extraction}

The prompt used for generating text segments is shown in Figure~\ref{fig:prompt}.

\figPrompt

\section{Motion Segmentation Module}

While equal segmentation provides a simple yet effective baseline, it does not explicitly capture the semantic structure of motion. Therefore, we further explored other segmentation methods, including the Change-Point-Detection (CPD)-based method that focuses on detecting motion transitions, and the clustering-based method that aims to detect motion primitives. Additionally, we process the BABEL~\cite{BABEL} annotations to create a validation set for evaluating these segmentation methods.

\paragraph{Change-Point-Detection (CPD)-based Segmentation:} 
The CPD-based method detects change points directly from a motion sequence and treats them as segmentation boundaries. We first apply a sliding window to extract motion segments from the complete motion sequence, and then employ a kernel-based change point detection method to identify change points within these segments. Specifically, we use a Gaussian kernel for detection, implemented using the Python \texttt{ruptures}~\cite{ruptures} package.

\paragraph{Clustering-based Segmentation:} 
The main idea of clustering-based segmentation is to first extract motion primitives from large-scale motion data and then utilize them to segment motion sequences. This method consists of two steps:

\textit{Motion Primitive Construction}: A motion primitive is defined as a short motion segment representing a specific motion pattern. Starting with a large motion dataset, such as AMASS~\cite{AMASS}, we apply a sliding window to extract a large number of motion segments. These segments are then grouped using a clustering algorithm such as K-Means~\cite{KMeans}, with similar motion segments clustered together to form a motion primitive library.

\textit{Motion Segmentation}: After constructing the motion primitive library, we use it to segment the motion sequences. Given a motion sequence, the goal is to segment it into distinct motion patterns. First, we apply a sliding window to extract all motion segments. For each segment, we compute its distance to the center of each motion primitive cluster, yielding a distance matrix $D \in \mathbb{R}^{N \times K}$, where $N$ is the number of segments and $K$ is the number of clusters. This distance matrix is treated as a cost matrix, and the optimal segmentation is determined by finding the cutting points that minimize the total distance, which can be efficiently solved using a dynamic programming algorithm.

\paragraph{Evaluation:} 
BABEL~\cite{BABEL} provides frame-level action labels for the AMASS dataset. However, a single frame may be associated with multiple labels, which is not suitable for our setting. To address this, we first process the annotations to ensure that each frame contains only one label, and then extract the segmentation points between consecutive segments. Based on this, we construct a validation set to evaluate the performance of different segmentation methods. We measure the error of the segmentation points in the token sequence, and the comparison is summarized in Table~\ref{tab:comp_seg}. The results show that the CPD-based and clustering-based methods yield a lower mean error, while uniform segmentation achieves a smaller variance of the errors.

\tabCompSeg

\section{Motion Aggregation Module}

The aggregation module aggregates motion tokens within a segment into a motion segment token, and we experimented with several strategies: (1) Mean Pooling, (2) Max Pooling, (3) Query-to-Token Cross-Attention, and (4) CLS-token Self-Attention. The comparison of different motion aggregation modules is presented in Figure~\ref{fig:agg_comp}. As illustrated in Table~\ref{tab:llm_agg_dis}, Mean-Max aggregation achieves the best overall performance.

\figAggComp

\section{Text-Motion Alignment Module}

Here we present more details of the alignment modules we have explored in the discussion section, including the Batch-Level Alignment, Global Alignment, and Token Alignment.

\paragraph{Batch-Level Alignment:} 
We introduce a batch-level alignment module, where the contrastive loss is computed across all segments within a batch, which extends the negative samples beyond those within a sample. The alignment loss is defined as follows:
\fmlBatchOne


\paragraph{Global Alignment:} 
Global alignment aligns the entire textual description with the complete motion sequence. The alignment loss is defined as follows:
\fmlGlobal
\noindent $\mathrm{sim}(T^i, M^j)$ denotes the cosine similarity between the $i$-th text and $j$-th motion sequence. 

\paragraph{Token Alignment:} 
Token alignment aligns each motion token with the text segment it belongs to. The alignment loss is defined as follows:
\fmlToken
\noindent $L$ denotes the maximum motion length, and $\mathbf{t}_{seg(\mathbf{x}_j^i)}^i$ denotes the corresponding text segment to which the motion token $\mathbf{x}_j^i$ belongs.

\paragraph{Comparison Analysis:} 
Although batch-level alignment introduces more negative samples, its performance does not surpass that of our proposed method. A possible reason is that similar or identical motion segments often exist within a batch—especially for common actions such as walking or running—making it difficult for the model to discriminate among them. Global alignment, on the other hand, ignores fine-grained segment correspondences and therefore fails to capture the temporal structure between text and motion. Token alignment fails to align a static motion token with a text segment representing a dynamic motion. 

\section{Segment Number Distribution Analysis}

Figure~\ref{fig:distribution} shows the segment number distributions of the HumanML3D~\cite{T2M} and KIT-ML~\cite{KIT-ML} datasets across training, validation, and test sets. More than half of the samples in HumanML3D contain multiple segments, whereas most samples in KIT-ML consist of only a single segment. This discrepancy highlights that richer compositional structure makes segment-level alignment more beneficial. Consequently, our alignment module has a more pronounced effect on HumanML3D compared to KIT-ML.

\figDistribution

\section{Failure Cases}

Figure~\ref{fig:failure} shows some failure cases of our method. In Figure~\ref{fig:failure} (a), although our method generates a marching motion while raising the arm, it neglects the keyword \textit{``in place"}, leading to an incorrect result. This suggests that our model sometimes struggles to capture subtle textual modifiers. In Figure~\ref{fig:failure} (b), when the specified motion length is short but the textual description contains multiple consecutive actions, our method may fail to generate the complete sequence, such as omitting the final \textit{``stop"} action. This suggests that stronger constraints are needed to encourage the model to allocate appropriate time durations for each action.

\figFailure


\end{document}